\documentclass[runningheads]{llncs}
\usepackage[T1]{fontenc}
\usepackage{graphicx}
\usepackage{url}
\usepackage{multirow}
\usepackage{booktabs} 
\usepackage{amssymb}
\usepackage{amsmath}
\usepackage[caption=false]{subfig}

\usepackage{color, colortbl}
\definecolor{Gray}{gray}{0.9}

\makeatletter
\newcommand{\printfnsymbol}[1]{%
  \textsuperscript{\@fnsymbol{#1}}%
}
\makeatother
\begin{document}
 \title{LifeLonger: A Benchmark for Continual Disease Classification }
\titlerunning{LifeLonger: A Benchmark for Continual Disease Classification}
%
\author{Mohammad Mahdi Derakhshani\inst{1}\thanks{Equal contribution} \and
Ivona Najdenkoska\inst{1}\printfnsymbol{1} \and
Tom van Sonsbeek\inst{1}\printfnsymbol{1} \and
Xiantong Zhen\inst{1,2} \and
Dwarikanath Mahapatra\inst{2} \and\\
Marcel Worring\inst{1} \and
Cees G. M. Snoek \inst{1}}

\authorrunning{M. Derakhshani et al.}
%
\institute{University of Amsterdam, Amsterdam, the Netherlands \and
Inception Institute of Artificial Intelligence, Abu Dhabi, UAE}
\maketitle 
\setcounter{footnote}{0} 
\begin{abstract}
Deep learning models have shown a great effectiveness in recognition of findings in medical images. However, they cannot handle the ever-changing clinical environment, bringing newly annotated medical data from different sources. To exploit the incoming streams of data, these models would  benefit largely from sequentially learning from new samples, without forgetting the previously obtained knowledge.
In this paper we introduce \textit{LifeLonger}, a benchmark for continual disease classification on the MedMNIST collection, by applying existing state-of-the-art continual learning methods. In particular, we consider three continual learning scenarios, namely, task and class incremental learning and the newly defined cross-domain incremental learning. Task and class incremental learning of diseases address the issue of classifying new samples without re-training the models from scratch, while cross-domain incremental learning addresses the issue of dealing with datasets originating from different institutions while retaining the previously obtained knowledge. We perform a thorough analysis of the performance and examine how the well-known challenges of continual learning, such as the catastrophic forgetting exhibit themselves in this setting. The encouraging results demonstrate that continual learning has a major potential to advance disease classification and to produce a more robust and efficient learning framework for clinical settings. The code repository, data partitions and baseline results for the complete benchmark are publicly available\footnote{\url{https://github.com/mmderakhshani/LifeLonger}}.

\keywords{Medical Continual Learning \and Disease Classification \and Medical Image Analysis.}
\end{abstract}

\section{Introduction}
Applying deep learning models to automate disease classification in medical images has a major potential to assist diagnosis in clinical practice and minimize labour \cite{LITJENS201760,liu2019comparison}. 
Current deep learning models achieve best performance when trained on large general datasets. 
However, adopting this to medical settings has additional challenges compared to when dealing with general data. 
Obtaining large enough datasets is difficult due to discrepancies in the imaging protocols and medical equipment across different hospitals. Additionally, learning from new incoming data requires constant retraining of existing models, which is less efficient.
An ideal approach would be to update models as new data arrives, without forgetting the knowledge obtained from previously seen datasets, which is the objective of this paper. 

In the machine learning parlance, the ability of a model to learn sequentially from incoming streams of data is known as \textit{continual} or \textit{lifelong learning} \cite{ring1998child,lopez2017gradient,nguyen2017variational}. Unlike the conventional supervised learning of tasks by observing the whole dataset and its label space at once, this sequential learning of tasks is constantly updating the knowledge of the model as it processes more data. Applying this learning paradigm would make deep learning models much more versatile to the constant growth of medical datasets. Therefore, training models for disease classification to learn sequentially is highly desired.
%
Despite the promise, very few efforts have been made to exploit continual learning in medical settings. Existing work tackles this paradigm in image segmentation \cite{baweja2018towards,gonzalez2020wrong,zheng2021continual,memmel2021adversarial,zhang2021comprehensive}, disease classification \cite{chakraborti2021contrastive,li2020continual,yang2021continual}, domain adaptation \cite{lenga2020continual} and domain incremental learning \cite{ge2021continual}, showing that they are only approaching a small spectrum of continual learning scenarios. They are bypassing more challenging scenarios, which are already considered for well-curated general imagery datasets and tasks \cite{lomonaco2021avalanche}. Consequently, no appropriate continual learning baselines exist to detect findings in medical images, creating additional challenges for this promising learning paradigm. 

This paper introduces the first continual learning benchmark on medical images for multi-class disease classification, by adopting five popular approaches, namely \textit{Elastic Weight Consolidation} (EWC) \cite{kirkpatrick2017overcoming}, \textit{Learning without Forgetting} (LwF)  \cite{li2017learning}, \textit{Memory Aware Synapses} (MAS)~\cite{zenke2017continual}, \textit{Incremental Classifier and Representation Learning} (iCaRL) \cite{Rebuffi2016iCaRLIC} and \textit{End-to-End Incremental Learning} (EEIL) \cite{castro2018end}. We train and evaluate models on the multi-class disease classification datasets of the publicly available MedMNIST \cite{medmnistv1}, which represents a large-scale MNIST-like collection of biomedical images. Furthermore, we analyse the performance by reporting the average accuracy and forgetting criteria. This analysis also represents an effort to address the major challenge in continual learning, which causes performance degradation on previous tasks after the model is trained on new tasks, named catastrophic forgetting \cite{goodfellow2013empirical,kirkpatrick2017overcoming,mccloskey1989catastrophic,li2017learning}.

In the scope of the benchmark, we introduce a new continual learning scenario, termed \textit{cross-domain incremental learning} where each dataset is treated as a distinct domain. This way of learning is especially practical when dealing with datasets originating from different hospitals or imaging equipment. Instead of being retrained from scratch for each specific dataset, models can benefit from sharing the learned knowledge across different datasets. Particularly, to better mimic a future clinical scenario, where a complete diagnosis of a patient is required beyond a single medical modality, our cross-domain incremental learning setting also assumes that disease classification tasks can come from different medical modalities. Moreover, the ability to aggregate knowledge from data coming from different institutions, without the need to re-train from scratch, provides another clear benefit for cross-domain incremental learning.
Secondly, we examine two types of existing continual learning scenarios, namely, \textit{task incremental} and \textit{class incremental learning}. In these two scenarios, the disease labels are grouped as separate subsets, representing the ``tasks''. In task incremental learning, the model is able to relate each data sample to one specific task \cite{masana2020class,van2019three} since it is aware of which classification label appears within each task. In the more complex class incremental learning setting, it is not known which task the data sample belongs to, making this type of learning considerably more difficult. These scenarios are highly useful in clinical practise, since the model should be able to respond well to new labels i.e. diseases in a dataset, while preserving the performance on previously seen labels and without being trained from scratch on it.

The following is the summary of our contributions: (1) We introduce the first benchmark on the MedMNIST dataset, for task, class and cross-domain incremental learning, as an effort to advance the continual learning methods in disease classification of medical images. (2) We introduce a new setting, termed cross-domain incremental learning for multi-class disease classification to illustrate a highly relevant clinical scenario of sharing learned knowledge across different medical datasets. (3) We explore task and class incremental learning scenarios of continual learning, to respond well to new labels i.e. diseases for multi-class disease classification.

\begin{figure}[t]
    \centering
    \includegraphics[width=\textwidth]{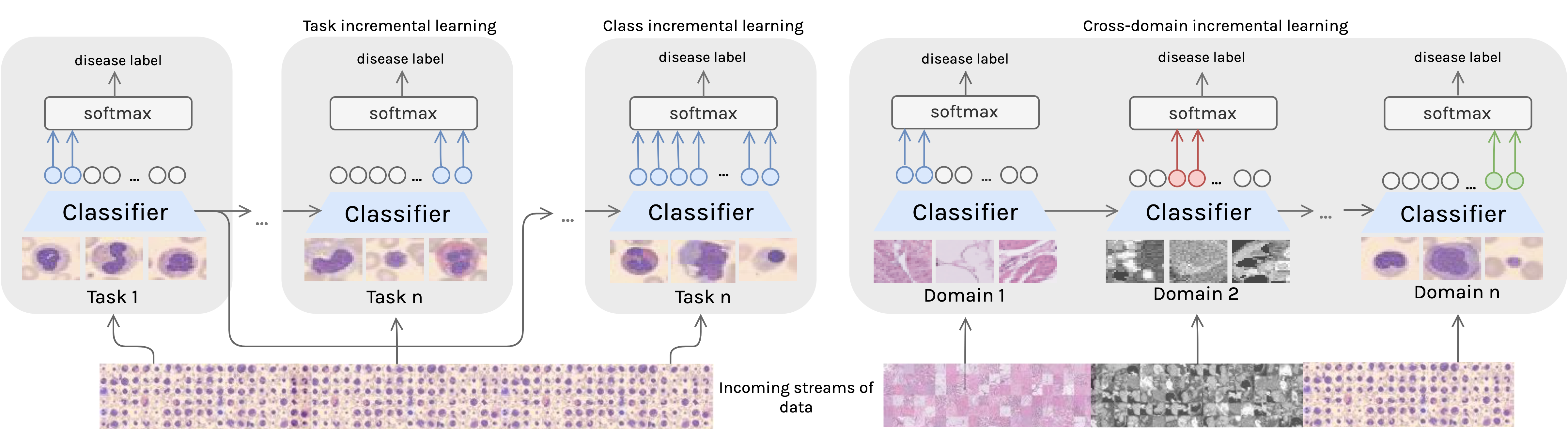}
    \caption{Continual learning scenarios covered in the LifeLonger benchmark: task incremental learning, class incremental learning and cross-domain incremental learning. Each scenario uses a random subset of the incoming data stream and its label space and each domain is a separate dataset (see Table \ref{tab:data}). The classifier is shared across all tasks and domains and it yields output logits (denoted by colored circles) representing the current task (and/or domain). We provide baseline results for all tasks.}
    \label{fig:task-class-cross-domain-il}
\end{figure}

\section{LifeLonger benchmark definition}
Formally, continual learning for multi-class disease classification consists of a sequence of $n$ non-stationary tasks $\{{\mathcal{T}_1, \mathcal{T}_2, \mathcal{T}_3, \cdots, \mathcal{T}_n}\}$, where each task $\mathcal{T}_i$ is represented by the training data $\mathcal{X}_i$ and its label space $\mathcal{Y}_i$, which refers to a random subset of disease labels. The tasks are disjoint, and there is no overlap between tasks' label space ($\mathcal{Y}_i \cap \mathcal{Y}_j = \emptyset$ for $i \neq j$). During the training of task $t$, the learner only has access to the corresponding subset $\mathcal{X}_t$. The model is trained once on each task using online incremental learning techniques or multiple times using offline learning approaches. In this paper, we utilize offline incremental learning.

\begin{table}[!t]
\centering
\caption{Multi-class disease datasets used in the LifeLonger benchmark, adopted from MedMNIST \cite{medmnistv1}. We follow the splits defined by MedMNIST, but add the class per task division to allow for continual learning.
%
}
\label{tab:benchmark_details}
\resizebox{0.8\columnwidth}{!}{%
\begin{tabular}{@{}lrrrrrc@{}}
\toprule
& \textbf{Train} & \textbf{Val} & \textbf{Test} & \textbf{Class} & \textbf{Tasks} & \textbf{Classes per Task} \\ 
\midrule
BloodMNISTs~\cite{acevedo2020dataset}  & 11,959 & 1,712 & 3,421 & 8  & 4 & [2, 2, 2, 2]  \\
OrganaMNIST~\cite{bilic2019liver}  & 34,581 & 6,491 & 17,778 & 11 & 4 & [3, 3, 3, 2]    \\
PathMNIST~\cite{kather2019predicting}   & 89,996 & 10,004 & 7,180 & 9  & 4 & [3, 2, 2, 2]    \\
TissueMNIST~\cite{bilic2019liver}  & 165,466 & 23,640 & 47,280 & 8  & 4 & [2, 2, 2, 2]    \\
\bottomrule
\end{tabular}%
\label{tab:data}
}
\end{table}

\subsection{Multi-class disease datasets}
We consider the MedMNIST collection \cite{medmnistv1} to define our benchmark, due to its balanced and standardized datasets spanning across various modalities. All images are normalized and rescaled to size $28{\times}28$ to enable fast computation and evaluation. We select subsets of the collection appropriate for multi-class disease classification, namely, BloodMNIST~\cite{acevedo2020dataset}, OrganaMNIST~\cite{bilic2019liver}, PathMNIST~\cite{kather2019predicting} and TissueMNIST~\cite{bilic2019liver}. 
We divide each dataset according to its label space $\mathcal{Y}$ into disjoint subsets ${(\mathcal{X}_1, \mathcal{Y}_1), (\mathcal{X}_2, \mathcal{Y}_2), \cdots, (\mathcal{X}_n, \mathcal{Y}_n)}$, such that $\mathcal{Y}_i \cap \mathcal{Y}_j {=} \emptyset$ for $i {\neq} j$, where each $(\mathcal{X}_i, \mathcal{Y}_i)$ is considered as a separate task.
Dataset details are provided in Table \ref{tab:data}.

\subsection{Continual learning scenarios}
We consider three continual learning scenarios to establish the benchmarks and to evaluate the model performance: task and class incremental learning~\cite{rusu2016progressive,derakhshani2021kernel,Rebuffi2016iCaRLIC} and the newly introduced cross-domain incremental learning.

The task incremental learning protocol uses knowledge of the task identifier $t$ and evaluates the model exclusively on the label space $\mathcal{Y}_t$. This is in contrast to class incremental learning, as a more challenging scenario in which the performance of the model is evaluated on all observed classes $\cup^t_{i=1}\mathcal{Y}_i$. 

The cross-domain incremental learning scenario allows to measure the ability of continual learning models in terms of transferring knowledge between different domains. In particular, each domain is defined as a separate dataset for multi-class disease classification. Formally, in this scenario, given a sequence of $n$ distinct domains, multi-class disease classification datasets, $\mathcal{D}_1, \mathcal{D}_2, \cdots \mathcal{D}_n$, we consider each domain $\mathcal{D}_i$ as a separate task $\mathcal{T}_i$. The learner is then trained on this series of tasks, same as in task incremental learning and class incremental learning scenarios. Additionally, we define \textit{domain-aware} and \textit{domain-agnostic} incremental learning. In domain-aware incremental learning, the domain identifier is available during inference, whereas in domain-agnostic incremental learning, this information does not exist.
Figure \ref{fig:task-class-cross-domain-il} illustrates these three continual learning scenarios.

\subsection{Evaluation criteria}
We quantify continual learning performance by examining average accuracy and average forgetting criterion, following existing work~\cite{Rebuffi2016iCaRLIC,derakhshani2021kernel}. When a training task $t$ is complete, the average accuracy of a model is computed as $\mathrm{A}_{t}{=}\frac{1}{t} \sum_{i=1}^{t} \mathrm{a}_{t, i}$,
where ${a}_{t, i}$ indicates the accuracy of the model on task $i$ when training on task $t$ is finished. Average forgetting quantifies the decline in the model performance between the highest and lowest accuracy for each task, calculated as $\mathrm{F}{=}\frac{1}{T-1} \sum_{i=1}^{T-1} \max _{1, \ldots, \mathrm{T}-1}\left(\mathrm{a}_{\mathrm{t}, \mathrm{i}}-\mathrm{a}_{\mathrm{T}, \mathrm{i}}\right)$.

\section{Baseline continual learners}
We provide implementation code for five state-of-the-art continual learners, covering all existing categories of continual learning methods, namely, regularization methods, rehearsal methods and bias correction methods. We establish a lower bound (LB) by simply fine-tuning the model on the current task, without relying on any specific continual learning strategy. Moreover, we provide the multi-task learning average accuracy for each benchmark as the upper bound (UB).

\textbf{Regularization methods} They reduce catastrophic forgetting by combining regularization term with the classification loss. While some algorithms such as \textit{Elastic Weight Consolidation} (EWC) and \textit{Memory Aware Synapses} (MAS) regularize the weights and estimate an importance measure for each parameter in the network~\cite{kirkpatrick2017overcoming,aljundi2018memory,zenke2017continual,chaudhry2018riemannian}, others, e.g. \textit{Learning without Forgetting} (LwF), regularize the feature map and try to minimize activation drift via knowledge distillation~\cite{li2017learning}. 

\textbf{Rehearsal-based methods} These methods assume the availability of data from previous tasks via a fixed-size memory unit~\cite{Rebuffi2016iCaRLIC,wu2019large,castro2018end}, a generative model capable of synthesizing samples~\cite{shin2017continual,ostapenko2019learning} or pseudo samples~\cite{kemker2018fearnet,xiang2019incremental} from previous tasks. Rehearsal systems aim to avoid forgetting by replaying previously stored or produced data from earlier tasks. From this category, we evaluate on \textit{Incremental Classifier and Representation Learning} (iCaRL)~\cite{Rebuffi2016iCaRLIC}.

\textbf{Bias correction methods} Regularization and rehearsal approaches are primarily affected by task recency bias, which refers to a network's inclination to be biased toward classes, associated with the most recently learnt task. This is partly because by the time the training is complete, the network has seen numerous instances of classes in the most recent task but none (or very few in the case of rehearsal) in the prior tasks. Methods for bias correction are meant to resolve this issue~\cite{hou2019learning,castro2018end,wu2019large}. A simple yet effective approach is proposed by Castro et al.~\cite{castro2018end}, termed \textit{End-to-End Incremental Learning} (EEIL), in which they suggest a balanced training step at the end of each training session. This phase uses an equal number of exemplars from each class for a specified number of iterations.

\subsection{Implementation details}
As the learner, we use a deep neural network parameterized by weights $\theta$, to transform the input data $x$ to the output logits $o$. We split this network into two parts: a feature extractor $h$ parameterized by $\psi$, and a classifier $f$ in particular a fully-connected layer with parameters $\phi$. To predict the label, we apply a softmax layer on the network logits $\hat{y}=\sigma(o)$ where $o=f_\phi(h_\psi(x))$.
For all baselines, we use a ResNet-18~\cite{he2016deep} as $h_\theta$ (without the penultimate fully-connected layers), trained on five distinct random seeds across four sequential tasks. $f_\phi$ includes a set of fully-connected layers with 512 neurons. To provide fair comparisons, we train the model using the same hyperparameters for all baselines.
All baseline runs have a batch size of $32$. For each task, we train the model on one NVIDIA RTX 2080ti GPU for 200 epochs with the option of early stopping in the occurrence of overfitting. 

\section{Baseline results}

\begin{figure}[t]
\centering
\includegraphics[width = \textwidth]{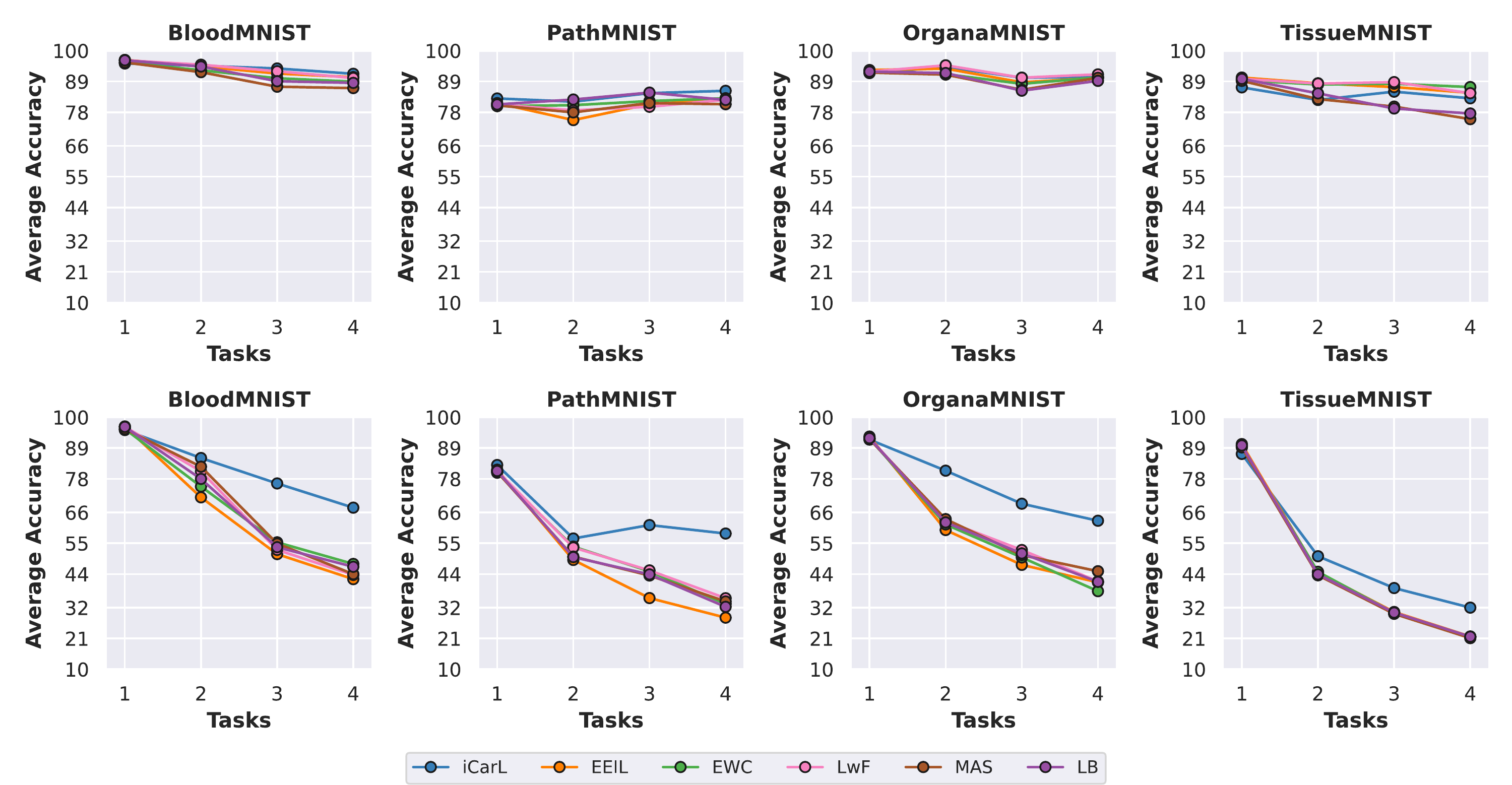}
\caption{LifeLonger benchmark results for four consecutive datasets in terms of average accuracy and average forgetting. The results show that all baselines for task incremental learning (top row) perform similarly, whereas for class incremental learning (bottom row) we observe better performance of iCarL, a conclusion that deviates from existing benchmarks for general imagery datasets \cite{masana2020class}. } 
\label{com:running_avg_forg_all}
\end{figure}

\begin{table*}[t]
\centering
\caption{
LifeLonger benchmark results for the provided baselines in terms of average accuracy and average forgetting.
All baselines perform comparably in task incremental learning, while iCarL consistently outperforms other alternatives in class incremental learning, suggesting the potential of the rehearsal-based approaches for disease classification.}
\resizebox{1\linewidth}{!}{%
\begin{tabular}{lcccccccc}
\toprule
\multicolumn{9}{c}{\textbf{Task Incremental Learning}}\\ \midrule
\multirow{2}{*}{\textbf{Method}} & 
\multicolumn{2}{c}{\textbf{BloodMNIST}} & 
\multicolumn{2}{c}{\textbf{PathMNIST}} & 
\multicolumn{2}{c}{\textbf{OrganaMNIST}} & 
\multicolumn{2}{c}{\textbf{TissueMNIST}} \\  \cmidrule(lr){2-3} \cmidrule(lr){4-5} \cmidrule(lr){6-7} \cmidrule(lr){8-9} 
 & Accuracy~$\uparrow$ & Forgetting~$\downarrow$ & Accuracy~$\uparrow$ & Forgetting~$\downarrow$ & Accuracy~$\uparrow$ & Forgetting~$\downarrow$ & Accuracy~$\uparrow$ & Forgetting~$\downarrow$\\ \midrule
LB  & 88.49\scriptsize{$\pm$4.09} & 9.84\scriptsize{$\pm$7.10}  & 82.40\scriptsize{$\pm$10.18} & 12.60\scriptsize{$\pm$19.80} & 89.21\scriptsize{$\pm$7.59} & 2.58\scriptsize{$\pm$4.75} & 77.56\scriptsize{ $\pm$10.61} & 12.1\scriptsize{$\pm$7.16} \\
EWC~\cite{kirkpatrick2017overcoming} & 88.94\scriptsize{$\pm$2.16} & 9.36\scriptsize{$\pm$6.12}  & 83.01\scriptsize{$\pm$3.89} & 6.48\scriptsize{$\pm$11.88} & 90.43\scriptsize{$\pm$3.94} & 3.07\scriptsize{$\pm$3.32} & \textbf{86.99\scriptsize{$\pm$3.71}} & \textbf{1.54\scriptsize{$\pm$0.94}} \\
MAS~\cite{zenke2017continual} & 86.63\scriptsize{$\pm$7.27} & 10.72\scriptsize{$\pm$6.02}  & 80.87\scriptsize{$\pm$5.63} & 4.64\scriptsize{$\pm$5.77} & 90.29\scriptsize{$\pm$5.12} & 2.41\scriptsize{$\pm$2.61} & 75.54\scriptsize{$\pm$9.82} & 5.16\scriptsize{$\pm$3.38} \\
LwF~\cite{li2017learning}  & 90.24\scriptsize{$\pm$4.04} & \textbf{0.97\scriptsize{$\pm$1.06}}   & 82.47\scriptsize{$\pm$8.26} & \textbf{-2.24\scriptsize{$\pm$4.88}} & \textbf{91.52\scriptsize{$\pm$5.50}} & -0.10\scriptsize{$\pm$0.96} & 84.81\scriptsize{$\pm$5.92} & 2.16\scriptsize{$\pm$3.51}\\
iCarL~\cite{Rebuffi2016iCaRLIC} & \textbf{91.74\scriptsize{$\pm$3.22}} &  1.83\scriptsize{$\pm$2.53} & \textbf{85.66\scriptsize{$\pm$7.95}}  & 0.28\scriptsize{$\pm$2.65} & 90.66\scriptsize{$\pm$7.21} & 2.27\scriptsize{$\pm$2.21} & 82.98\scriptsize{$\pm$5.67} & 2.97\scriptsize{$\pm$2.31}\\
EEIL~\cite{castro2018end} & 90.55\scriptsize{$\pm$2.90} & 2.01\scriptsize{$\pm$2.14}   & 82.65\scriptsize{$\pm$5.58} & -0.24\scriptsize{$\pm$6.18} & 90.30\scriptsize{$\pm$8.68} & \textbf{-0.27\scriptsize{$\pm$3.14}} & 84.95\scriptsize{$\pm$4.56} & 3.77\scriptsize{$\pm$4.36}\\
\rowcolor{Gray}
UB & 97.98\scriptsize{$\pm$0.18} & - & 93.52\scriptsize{$\pm$1.91} & - & 95.22\scriptsize{$\pm$0.37} & - & 91.27\scriptsize{$\pm$0.87} & - \\
\midrule
\midrule
\multicolumn{9}{c}{\textbf{Class Incremental Learning}}\\
\midrule
\multirow{2}{*}{\textbf{Method}} & 
\multicolumn{2}{c}{\textbf{BloodMNIST}} & 
\multicolumn{2}{c}{\textbf{PathMNIST}} & 
\multicolumn{2}{c}{\textbf{OrganaMNIST}} & 
\multicolumn{2}{c}{\textbf{TissueMNIST}} \\  \cmidrule(lr){2-3} \cmidrule(lr){4-5} \cmidrule(lr){6-7} \cmidrule(lr){8-9} 
 & Accuracy~$\uparrow$ & Forgetting~$\downarrow$ & Accuracy~$\uparrow$ & Forgetting~$\downarrow$ & Accuracy~$\uparrow$ & Forgetting~$\downarrow$ & Accuracy~$\uparrow$ & Forgetting~$\downarrow$\\
\midrule
LB  & 46.59\scriptsize{$\pm$6.46} & 68.26\scriptsize{$\pm$8.13}  & 32.29\scriptsize{$\pm$6.74} & 77.54\scriptsize{$\pm$16.40} & 41.21\scriptsize{$\pm$7.39} & 54.20\scriptsize{$\pm$23.97} & 21.63\scriptsize{ $\pm$2.55} & 90.69\scriptsize{$\pm$1.77} \\
EWC~\cite{kirkpatrick2017overcoming} & 47.60\scriptsize{$\pm$7.30} & 66.22\scriptsize{$\pm$14.36}  & 33.34\scriptsize{$\pm$4.77} & 76.39\scriptsize{$\pm$15.53} & 37.88\scriptsize{$\pm$7.97} & 67.92\scriptsize{$\pm$27.40} & 21.56\scriptsize{$\pm$2.51} & 90.48\scriptsize{$\pm$1.76} \\
MAS~\cite{zenke2017continual} & 43.94\scriptsize{$\pm$6.96} & 69.43\scriptsize{$\pm$12.90}  & 34.22\scriptsize{$\pm$6.17} & 74.98\scriptsize{$\pm$15.79} & 44.99\scriptsize{$\pm$5.61} & 55.13\scriptsize{$\pm$24.43} & 21.11\scriptsize{$\pm$2.56} & 89.38\scriptsize{$\pm$1.72} \\
LwF~\cite{li2017learning}  & 43.68\scriptsize{$\pm$6.55} & 66.30\scriptsize{$\pm$8.70}   & 35.36\scriptsize{$\pm$9.59} & 67.37\scriptsize{$\pm$15.76} & 41.36\scriptsize{$\pm$6.36} & 51.47\scriptsize{$\pm$24.53} & 21.49\scriptsize{$\pm$2.58} & 90.79\scriptsize{$\pm$1.22}\\
iCarL~\cite{Rebuffi2016iCaRLIC} & \textbf{67.70\scriptsize{$\pm$8.67}} & \textbf{14.52\scriptsize{$\pm$6.93}} & \textbf{58.46\scriptsize{$\pm$8.79}} & \textbf{-0.70\scriptsize{$\pm$6.41}} & \textbf{63.02\scriptsize{$\pm$7.53}} & \textbf{7.75\scriptsize{$\pm$4.49}} & \textbf{32.00\scriptsize{$\pm$3.01}} & \textbf{14.42\scriptsize{$\pm$10.21}} \\
EEIL~\cite{castro2018end} & 42.17\scriptsize{$\pm$7.05} & 71.25\scriptsize{$\pm$12.90}   & 28.42\scriptsize{$\pm$5.43} & 79.39\scriptsize{$\pm$15.79} & 41.03\scriptsize{$\pm$11.53} & 62.47\scriptsize{$\pm$24.43} & 21.69\scriptsize{$\pm$2.43} & 91.35\scriptsize{$\pm$1.72}\\
\rowcolor{Gray}
UB & 97.98\scriptsize{$\pm$0.18} & - & 93.52\scriptsize{$\pm$1.91} & - & 95.22\scriptsize{$\pm$0.37} & - & 91.27\scriptsize{$\pm$0.87} & - \\
\bottomrule
\end{tabular}%
}
\label{tab:compare-TIL}
\end{table*}

\begin{table*}[!htbp]
\centering
\caption{LifeLonger benchmark results for the provided baselines on the domain-aware and domain-agnostic incremental learning tasks, in terms of average accuracy and average forgetting, where each domain is represented by a separate dataset.
iCarL consistently outperforms alternatives in both cases, suggesting again the potential of the rehearsal-based approaches for continual disease classification. }
\resizebox{0.6\linewidth}{!}{%
\begin{tabular}{lcccc}
\toprule
& \multicolumn{2}{c}{\textbf{Domain-aware}} & 
\multicolumn{2}{c}{\textbf{Domain-agnostic}}  \\  \cmidrule(lr){2-3} \cmidrule(lr){4-5}  
 \textbf{Baselines} & Accuracy~$\uparrow$ & Forgetting~$\downarrow$ & Accuracy~$\uparrow$ & Forgetting~$\downarrow$ \\ \midrule
EWC~\cite{kirkpatrick2017overcoming} &   28.61\scriptsize{$\pm$4.99} &           50.49\scriptsize{$\pm$1.95}  & 21.59\scriptsize{$\pm$5.33} &          58.34\scriptsize{$\pm$5.86}  \\
LwF~\cite{li2017learning}            &   40.95\scriptsize{$\pm$3.98} &           42.58\scriptsize{$\pm$9.88}   & 37.95\scriptsize{$\pm$5.61} & 52.92\scriptsize{$\pm$11.89} \\
iCarL~\cite{Rebuffi2016iCaRLIC}& \textbf{51.22\scriptsize{$\pm$1.49}} &  \textbf{22.00\scriptsize{$\pm$0.75}} & \textbf{50.78\scriptsize{$\pm$1.51}}  & \textbf{21.78\scriptsize{$\pm$1.29}} \\
\rowcolor{Gray}
UB & 93.28\scriptsize{$\pm$0.28} & - & 93.28\scriptsize{$\pm$0.28} & -\\
\bottomrule
\end{tabular}%
}
\label{tab:compare-cross-domain}
\end{table*}

\subsubsection{Task and Class incremental learning} To compare the various baselines, we provide the average accuracy and forgetting, for four sequential tasks on five distinct random seeds among four medical datasets, given in Table \ref{tab:compare-TIL}. In terms of accuracy, all benchmarks show comparable performance, whereas in terms of the forgetting it can be noticed a larger gap between some approaches, indicating the drop of performance over time. Additionally, in Figure \ref{com:running_avg_forg_all} we illustrate the running average accuracy to show the performance over time. In task incremental learning, all baselines exhibit comparable performance, whereas, in class incremental learning the iCarL method highlights its great superiority and effectiveness over other alternatives across all benchmarks by a large margin. This is in contrast with the findings of previous work \cite{masana2020class}, showing that the EEIL method performs remarkably better than iCaRL in task and class incremental learning. Hence, this demonstrates the superiority of the rehearsal-based strategies for disease classification and indicates its potential for future study.

\subsubsection{Cross-domain incremental learning} 
\begin{figure}[t]
\centering
  \includegraphics[clip,width=0.95\columnwidth]{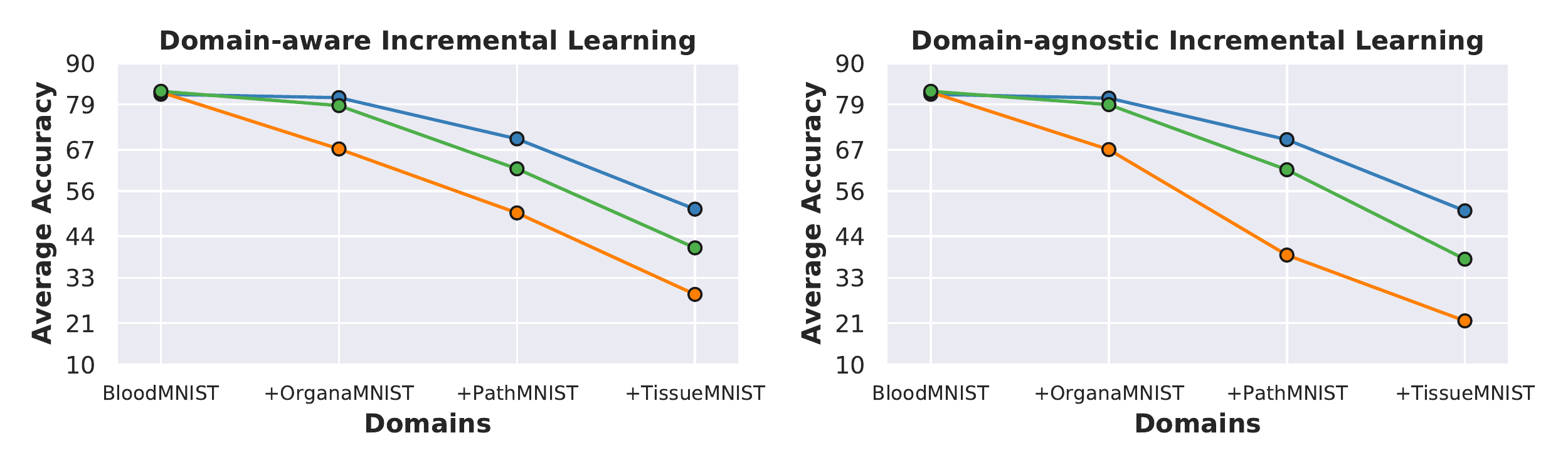}%
\vfill
  \includegraphics[clip, trim=4.5cm 0cm 0cm 0cm, width=0.95\textwidth]{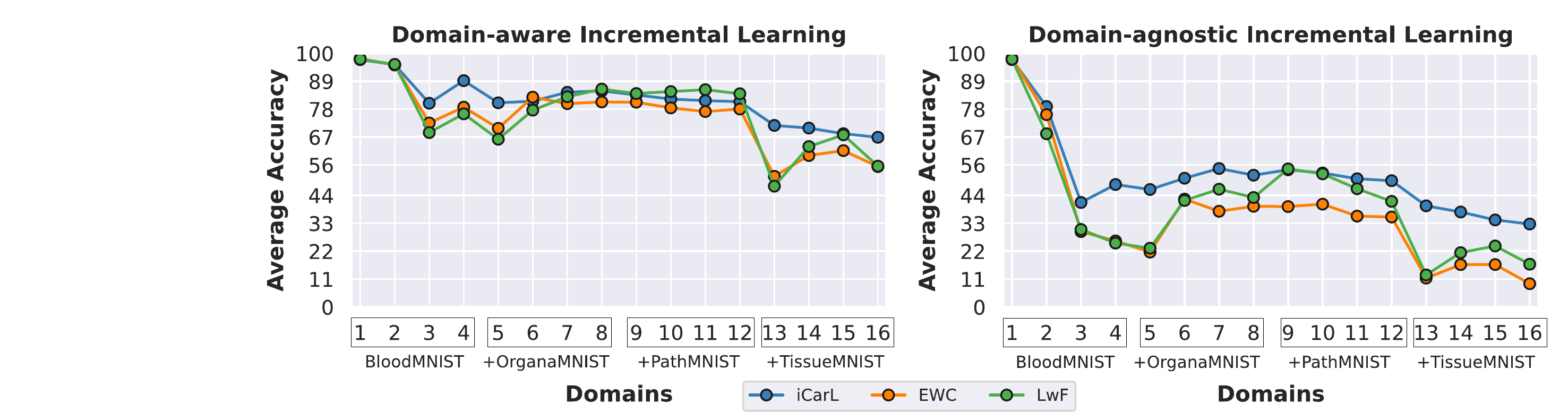}%
\caption{LifeLonger benchmark results for cross-domain incremental learning. \textit{Left} column presents the average accuracy of domain-aware incremental learning while \textit{right} column indicates the average forgetting of  domain-agnostic incremental learning. 
}
\label{fig:comp-cross-domain}
\end{figure}

We conduct two separate experiments to evaluate the performance for disease classification of domain-aware and domain-agnostic incremental learning. For the first experiment, we report the average accuracy and forgetting on the best performing approaches, namely, EWC, LwF and iCarL. We treat each dataset (domain) as a separate task and train a continual learning model and we report the results on Table \ref{tab:compare-cross-domain}. It can be observed that iCarL constantly outperforms LwF and EWC, once again proving its superiority for disease classification.
Similarly as in the previous scenarios, we report the average accuracy across time of domain-aware and domain-agnostic models, in the top row of Figure \ref{fig:comp-cross-domain}. This shows that both scenarios are similar in terms of complexity when we treat each domain as a separate task.
The second experiment ablates a fine-grained version of cross-domain incremental learning by subdividing each domain into a sequence of tasks, for example, four tasks per domain. The bottom row of Figure \ref{fig:comp-cross-domain} shows the running average accuracy for domain-aware and domain-agnostic classification. As shown, iCarL still outperforms other approaches. Additionally, these experiments suggest that fine-grained cross-domain incremental learning improves the performance of domain-aware incremental learning, while it deteriorates that of domain-agnostic incremental learning, suggesting opportunities for improvements of existing approaches.

\section{Conclusion}
In this paper we introduce the first benchmark for continual learning for disease classification of medical images. We introduce cross-domain incremental learning which shows to be a suited approach when dealing with datasets originating from different institutions.
Additionally, we adopt task and class incremental learning to illustrate a relevant clinical scenario when the model should readjust to newly labeled samples without being trained from scratch on previously seen data. Last, but not least, we evaluate various state-of-the-art methods, showing that the rehearsal-based methods are the most promising category of methods for disease classification.
The obtained results demonstrate the shortcomings of the current continual learning methods for disease classification of medical images, due to their inherent complexity, such as the spatial locality of diseases, compared to general images. Nevertheless, this paper represents an initial effort to establish the foundation of continual learning for disease classification of medical images.

\section*{Acknowledgements}
This work is financially supported by the Inception Institute of Artificial Intelligence, the University of Amsterdam and the allowance Top consortia for Knowledge and Innovation (TKIs) from the Netherlands Ministry of Economic Affairs and Climate Policy.
%
%
%
\bibliographystyle{splncs04}
\bibliography{references}

\begin{thebibliography}{10}
\providecommand{\url}[1]{\texttt{#1}}
\providecommand{\urlprefix}{URL }
\providecommand{\doi}[1]{https://doi.org/#1}

\bibitem{acevedo2020dataset}
Acevedo, A., Merino, A., Alf{\'e}rez, S., Molina, {\'A}., Bold{\'u}, L.,
  Rodellar, J.: A dataset of microscopic peripheral blood cell images for
  development of automatic recognition systems  (2020)

\bibitem{aljundi2018memory}
Aljundi, R., Babiloni, F., Elhoseiny, M., Rohrbach, M., Tuytelaars, T.: Memory
  aware synapses: Learning what (not) to forget. In: ECCV (2018)

\bibitem{baweja2018towards}
Baweja, C., Glocker, B., Kamnitsas, K.: Towards continual learning in medical
  imaging. arXiv preprint arXiv:1811.02496  (2018)

\bibitem{bilic2019liver}
Bilic, P., Christ, P., Vorontsov, E., Chlebus, G., Chen, H., Dou, Q., Fu, C.,
  Han, X., Heng, P., Hesser, J., et~al.: The liver tumor segmentation benchmark
  (lits). arxiv 2019. ArXiv  (2019)

\bibitem{castro2018end}
Castro, F.M., Mar{\'\i}n-Jim{\'e}nez, M.J., Guil, N., Schmid, C., Alahari, K.:
  End-to-end incremental learning. In: ECCV (2018)

\bibitem{chakraborti2021contrastive}
Chakraborti, T., Gleeson, F., Rittscher, J.: Contrastive representations for
  continual learning of fine-grained histology images. In: International
  Workshop on Machine Learning in Medical Imaging (2021)

\bibitem{chaudhry2018riemannian}
Chaudhry, A., Dokania, P.K., Ajanthan, T., Torr, P.H.: Riemannian walk for
  incremental learning: Understanding forgetting and intransigence. In: ECCV
  (2018)

\bibitem{derakhshani2021kernel}
Derakhshani, M.M., Zhen, X., Shao, L., Snoek, C.: Kernel continual learning.
  In: ICML (2021)

\bibitem{gonzalez2020wrong}
Gonzalez, C., Sakas, G., Mukhopadhyay, A.: What is wrong with continual
  learning in medical image segmentation? ArXiv  (2020)

\bibitem{goodfellow2013empirical}
Goodfellow, I.J., Mirza, M., Xiao, D., Courville, A., Bengio, Y.: An empirical
  investigation of catastrophic forgetting in gradient-based neural networks.
  ArXiv  (2013)

\bibitem{he2016deep}
He, K., Zhang, X., Ren, S., Sun, J.: Deep residual learning for image
  recognition. In: CVPR (2016)

\bibitem{hou2019learning}
Hou, S., Pan, X., Loy, C.C., Wang, Z., Lin, D.: Learning a unified classifier
  incrementally via rebalancing. In: CVPR (2019)

\bibitem{kather2019predicting}
Kather, J.N., Krisam, J., Charoentong, P., Luedde, T., Herpel, E., Weis, C.A.,
  Gaiser, T., Marx, A., Valous, N.A., Ferber, D., et~al.: Predicting survival
  from colorectal cancer histology slides using deep learning: A retrospective
  multicenter study  (2019)

\bibitem{kemker2018fearnet}
Kemker, R., Kanan, C.: Fearnet: Brain-inspired model for incremental learning.
  In: ICLR (2018)

\bibitem{kirkpatrick2017overcoming}
Kirkpatrick, J., Pascanu, R., Rabinowitz, N., Veness, J., Desjardins, G., Rusu,
  A.A., Milan, K., Quan, J., Ramalho, T., Grabska-Barwinska, A., Hassabis, D.,
  Clopath, C., Kumaran, D., Hadsell, R.: Overcoming catastrophic forgetting in
  neural networks. Proceedings of the National Academy of Sciences  (2017)

\bibitem{lenga2020continual}
Lenga, M., Schulz, H., Saalbach, A.: Continual learning for domain adaptation
  in chest x-ray classification. In: Medical Imaging with Deep Learning (2020)

\bibitem{li2017learning}
Li, Z., Hoiem, D.: Learning without forgetting. PAMI  (2017)

\bibitem{li2020continual}
Li, Z., Zhong, C., Wang, R., Zheng, W.S.: Continual learning of new diseases
  with dual distillation and ensemble strategy. In: MICCAI. pp. 169--178.
  Springer (2020)

\bibitem{LITJENS201760}
Litjens, G., Kooi, T., Bejnordi, B.E., Setio, A.A.A., Ciompi, F., Ghafoorian,
  M., van~der Laak, J.A., van Ginneken, B., Sánchez, C.I.: A survey on deep
  learning in medical image analysis. Medical Image Analysis  (2017)

\bibitem{liu2019comparison}
Liu, X., Faes, L., Kale, A.U., Wagner, S.K., Fu, D.J., Bruynseels, A.,
  Mahendiran, T., Moraes, G., Shamdas, M., Kern, C., et~al.: A comparison of
  deep learning performance against health-care professionals in detecting
  diseases from medical imaging: a systematic review and meta-analysis. The
  lancet digital health  (2019)

\bibitem{lomonaco2021avalanche}
Lomonaco, V., Pellegrini, L., Cossu, A., Carta, A., Graffieti, G., Hayes, T.L.,
  De~Lange, M., Masana, M., Pomponi, J., Van~de Ven, G.M., et~al.: Avalanche:
  an end-to-end library for continual learning. In: CVPR (2021)

\bibitem{lopez2017gradient}
Lopez-Paz, D., Ranzato, M.: Gradient episodic memory for continual learning.
  In: NeurIPS (2017)

\bibitem{masana2020class}
Masana, M., Liu, X., Twardowski, B., Menta, M., Bagdanov, A.D., van~de Weijer,
  J.: Class-incremental learning: survey and performance evaluation on image
  classification. ArXiv  (2020)

\bibitem{mccloskey1989catastrophic}
McCloskey, M., Cohen, N.J.: Catastrophic interference in connectionist
  networks: The sequential learning problem. In: Psychology of learning and
  motivation (1989)

\bibitem{memmel2021adversarial}
Memmel, M., Gonzalez, C., Mukhopadhyay, A.: Adversarial continual learning for
  multi-domain hippocampal segmentation. In: Domain Adaptation and
  Representation Transfer, and Affordable Healthcare and AI for Resource
  Diverse Global Health (2021)

\bibitem{nguyen2017variational}
Nguyen, C.V., Li, Y., Bui, T.D., Turner, R.E.: Variational continual learning.
  In: ICLR (2018)

\bibitem{ostapenko2019learning}
Ostapenko, O., Puscas, M., Klein, T., Jahnichen, P., Nabi, M.: Learning to
  remember: A synaptic plasticity driven framework for continual learning. In:
  CVPR (2019)

\bibitem{Rebuffi2016iCaRLIC}
Rebuffi, S.A., Kolesnikov, A.I., Sperl, G., Lampert, C.H.: {iCaRL}: Incremental
  classifier and representation learning. In: CVPR (2017)

\bibitem{ring1998child}
Ring, M.B.: Child: A first step towards continual learning. Learning to learn
  (1998)

\bibitem{rusu2016progressive}
Rusu, A.A., Rabinowitz, N.C., Desjardins, G., Soyer, H., Kirkpatrick, J.,
  Kavukcuoglu, K., Pascanu, R., Hadsell, R.: Progressive neural networks. In:
  NeurIPS (2016)

\bibitem{shin2017continual}
Shin, H., Lee, J.K., Kim, J., Kim, J.: Continual learning with deep generative
  replay. NeurIPS  (2017)

\bibitem{ge2021continual}
Srivastava, S., Yaqub, M., Nandakumar, K., Ge, Z., Mahapatra, D.: Continual
  domain incremental learning for chest x-ray classification in low-resource
  clinical settings. In: Domain Adaptation and Representation Transfer, and
  Affordable Healthcare and AI for Resource Diverse Global Health (2021)

\bibitem{van2019three}
Van~de Ven, G.M., Tolias, A.S.: Three scenarios for continual learning. ArXiv
  (2019)

\bibitem{wu2019large}
Wu, Y., Chen, Y., Wang, L., Ye, Y., Liu, Z., Guo, Y., Fu, Y.: Large scale
  incremental learning. In: CVPR (2019)

\bibitem{xiang2019incremental}
Xiang, Y., Fu, Y., Ji, P., Huang, H.: Incremental learning using conditional
  adversarial networks. In: CVPR (2019)

\bibitem{medmnistv1}
Yang, J., Shi, R., Ni, B.: Medmnist classification decathlon: A lightweight
  automl benchmark for medical image analysis. In: ISBI (2021)

\bibitem{yang2021continual}
Yang, Y., Cui, Z., Xu, J., Zhong, C., Wang, R., Zheng, W.S.: Continual learning
  with bayesian model based on a fixed pre-trained feature extractor. In:
  MICCAI. pp. 397--406. Springer (2021)

\bibitem{zenke2017continual}
Zenke, F., Poole, B., Ganguli, S.: Continual learning through synaptic
  intelligence. In: ICML (2017)

\bibitem{zhang2021comprehensive}
Zhang, J., Gu, R., Wang, G., Gu, L.: Comprehensive importance-based selective
  regularization for continual segmentation across multiple sites. In: MICCAI.
  pp. 389--399. Springer (2021)

\bibitem{zheng2021continual}
Zheng, E., Yu, Q., Li, R., Shi, P., Haake, A.: A continual learning framework
  for uncertainty-aware interactive image segmentation. In: AAAI (2021)

\end{thebibliography}
\end{document}